\documentclass[11pt]{article}

\usepackage[final]{acl}

\usepackage{times}
\usepackage{latexsym}

\usepackage[T1]{fontenc}

\usepackage[utf8]{inputenc}

\usepackage{microtype}

\usepackage{inconsolata}

\usepackage{amsmath}

\usepackage{enumitem}

\usepackage{graphicx}

\usepackage{listings}
\usepackage{tikz}
\usetikzlibrary{shapes.geometric, arrows.meta, positioning, fit, backgrounds}

\usepackage{booktabs}

\usepackage{pgfplots}
\pgfplotsset{compat=1.18}

\title{Shared Selective Persistent Memory for Agentic LLM Systems}

\author{Sanjana Pedada \\
  Apple Inc. \\
  \texttt{s\_pedada@apple.com} \And
  Aditya Dhavala \\
  Apple Inc. \\
  \texttt{a\_dhavala@apple.com} \And
  Neelraj Patil \\
  Apple Inc. \\
  \texttt{neelraj\_patil@apple.com}}

\begin{document}

\maketitle

\begin{abstract}
Agentic LLM systems that generate code through multi-turn tool use face a fundamental context problem: each session starts from zero, discarding the domain constraints, data schemas, tool configurations, and output preferences that made previous sessions productive. We introduce \textbf{shared selective persistent memory}, an architecture that retains four categories of \emph{reusable context}---task specifications, data schemas, tool configurations, and output constraints---while discarding session-specific reasoning traces, and that packages them into workspaces transferable across users under role-based access control.

The resulting cost curve is non-monotonic. In a controlled replication on four public datasets, where a formatting specification is established once and then withheld, no memory completes 0/12 trials at 3.8K input tokens, selective memory completes 12/12 at 3.9K, and full conversation history completes 8/12 at 7.7K. What is kept matters more than how much is kept: the winning configuration costs essentially what the failing one does, and twice as much context does not improve on it. Both differences from no memory survive Bonferroni-corrected exact McNemar tests ($p = 0.0005$, $p = 0.008$); selective memory and full history are indistinguishable on completion ($p = 0.125$) and separate on price. Every no-memory failure is a specification failure and nothing else---the artifacts are structurally sound but do not reproduce a constraint they were never given.

We implement this in a deployed platform where agents produce git-versioned artifacts from CSV, SQL, REST, and MCP sources. A complementary \textbf{zero-token data refresh} contract decouples generated programs from runtime data, firing on 12/12 trials at a median 0.08s with no model call, while summary-driven data representation costs 97--431$\times$ fewer tokens than raw injection. Across 24 recurring enterprise tasks selective memory completes 23/24 against 19/24 and 17/24, though at that sample no pairwise difference reaches significance; there, where users re-specify rather than rely on memory, full history returns no improvement at all despite 9$\times$ the input.
\end{abstract}

\section{Introduction}
\label{sec:intro}

Agentic LLM systems---those that autonomously invoke tools, write files, and execute code to accomplish user goals---have demonstrated impressive capabilities in code generation \citep{copilot2021, claude-code-2025}, data analysis \citep{openai-code-interpreter-2023}, and multi-step reasoning \citep{wu2023autogen}. Yet a fundamental limitation persists across all current agentic frameworks: \textbf{sessions are stateless}. When a conversation ends, the accumulated context---domain constraints, data schemas, tool configurations, prompt refinements, and output format preferences---is discarded entirely.

This is wasteful. Enterprise users repeatedly perform structurally similar tasks---weekly dashboards from updated exports, reports with consistent formatting, documents from evolving sources---and each session forces them to re-specify what the system should already know. The problem is not that LLMs lack memory, but that existing systems provide no mechanism to \emph{selectively persist} the context that matters while discarding the session-specific reasoning that does not.

The naive solution---persisting entire conversation histories---carries the context but does not pay for itself. A raw transcript does supply what a later session needs, lifting public-data completion from 0/12 to 8/12, but at twice the input tokens of a compact specification that reaches 12/12; on enterprise tasks, where the user re-specifies anyway, it returns no improvement at all despite 9$\times$ the input. We conjecture that a prior session's tool-use trace is not merely inert but misleading \citep{liu2024lost}, though our experiments bear out the cost rather than the mechanism.

We propose \textbf{shared selective persistent memory}, a memory architecture for agentic LLM systems built on a key observation: the reusable knowledge from an agentic session can be decomposed into four orthogonal categories, each of which is compact, stable across sessions, and beneficial when persisted:

\begin{enumerate}
    \item \textbf{Task specifications} --- custom system prompts that encode domain rules, output format constraints, and generation preferences (e.g., ``always use green for $\geq$95\% attainment'').
    \item \textbf{Data schemas} --- column names, types, statistical summaries, and relationships across data sources, precomputed from raw data.
    \item \textbf{Tool configurations} --- which external tools are available, their parameter schemas, invocation patterns, and authentication requirements.
    \item \textbf{Output constraints} --- structural contracts between the generated artifact and its runtime environment (e.g., ``data is injected at runtime, never hardcoded'').
\end{enumerate}

Crucially, what we \emph{discard} is equally important: the multi-turn reasoning trace, tool invocation logs, intermediate file states, and error-recovery paths from prior sessions. Our evidence is that these are not worth their price rather than that they are actively harmful---carrying them forward does supply a later session with the constraints it needs, but at roughly twice the input tokens of the four categories above, and with no measured gain.

We validate this architecture in a deployed platform where LLM agents produce and maintain git-versioned artifacts---dashboards, reports, data-driven documents---from heterogeneous sources (CSV, SQL, REST, MCP). A complementary \textbf{zero-token data refresh} mechanism separates the generated program from runtime-bound data, so data updates cost no LLM tokens.

Our contributions are:
\begin{enumerate}
    \item A \textbf{shared selective persistent memory architecture} that identifies four categories of reusable agentic context, persists them independently of session traces, and includes a \textbf{zero-token data refresh} mechanism that decouples generated programs from runtime data (Section~\ref{sec:memory}).
    \item A \textbf{collaborative workspace platform} with git-versioned artifacts, draft isolation, multi-connector data integration (CSV, SQL, REST, MCP), and AI-assisted editing and querying modes (Section~\ref{sec:sharing}).
    \item \textbf{Deployment case studies} across three enterprise scenarios demonstrating that shared selective memory reduces redundant specification, enables workspace reuse, and supports zero-token data refresh (Section~\ref{sec:evaluation}).
\end{enumerate}

\subsection{Motivating Use Cases}
\label{sec:usecases}

We ground our architecture in three enterprise scenarios. \textbf{(1)~Recurring refresh:} an analyst regenerates a supply chain dashboard weekly, re-specifying 8--12 formatting constraints each session over 4--5 turns; with selective memory those persist and a zero-token data swap refreshes the artifact directly. \textbf{(2)~Cross-team sharing:} a director publishes a revenue workspace as a versioned template, and a counterpart loads it, connects schema-compatible data, and sees the artifact render under role-based access control. \textbf{(3)~Iterative construction:} a team extends one workspace across sessions, with accumulated specifications carried forward, draft isolation protecting the published version, and git-backed revert restoring any prior state without re-invoking the model.

\section{Related Work}
\label{sec:related}

\paragraph{Memory in LLM Systems.}
Context window extension \citep{press2022train, chen2023extending} increases the token budget without addressing what to persist. RAG \citep{lewis2020retrieval} retrieves at the document level, not the structured configuration level agentic systems require. Conversation summarization \citep{xu2023beyond} compresses prior turns but retains session-specific reasoning traces. MemGPT \citep{packer2023memgpt} provides an OS-inspired memory hierarchy focused on conversational continuity. Generative Agents \citep{park2023generative} maintain memory streams optimized for narrative coherence. Reflexion \citep{shinn2023reflexion} persists failure summaries for within-task trial-and-error learning. Our approach identifies four structured categories to persist across sessions and explicitly discards reasoning traces.

\paragraph{Structured Memory and Context Compression.} A more recent line organizes agent memory as structured records rather than a transcript: A-MEM \citep{xu2025amem} builds an evolving note network, and Mem0 \citep{chhikara2025mem0} consolidates salient facts into a long-term store. LangGraph \citep{langgraph2024} separates a per-thread checkpointer from a cross-thread store, close in spirit to our session/workspace split. These systems learn what to retain, whereas our categories are fixed in advance and tied to the artifact-generation contract---less general, but directly actionable. A parallel line compresses agent context instead of restructuring memory \citep{kang2026acon, xiao2026agentdiet}; our selective forgetting sits at the coarse, declarative end of that space, discarding whole classes of session state by construction rather than scoring spans for retention.

\paragraph{Agentic LLM Frameworks.}
ReAct \citep{yao2023react} established interleaved reasoning and tool-use actions, building on chain-of-thought prompting \citep{wei2022chain}. LangChain \citep{langchain2022}, AutoGen \citep{wu2023autogen}, and CrewAI \citep{crewai2024} provide multi-agent orchestration. LATS \citep{zhou2023lats} unifies reasoning, acting, and planning via tree search. Voyager \citep{wang2023voyager} introduces a persistent skill library for embodied agents---the closest prior work, though it persists executable skills rather than declarative configuration. None of these frameworks persist task specifications, tool configurations, or output constraints across sessions.

\paragraph{Tool-Augmented LLMs.}
Toolformer \citep{schick2023toolformer} demonstrated autonomous tool learning; Gorilla \citep{patil2023gorilla} improved API call accuracy; MCP \citep{anthropic-mcp-2024} standardizes tool discovery. The OpenAI Assistants API \citep{openai-assistants-2023} provides thread-level persistence but retains full conversation threads rather than selectively extracting reusable configuration. LLM-powered code generation tools \citep{copilot2021, claude-code-2025, cursor2024} and code interpreters \citep{openai-code-interpreter-2023} are single-user and ephemeral. Traditional BI platforms provide persistent dashboards but require domain-specific query languages rather than natural language.

\section{Shared Selective Persistent Memory}
\label{sec:memory}

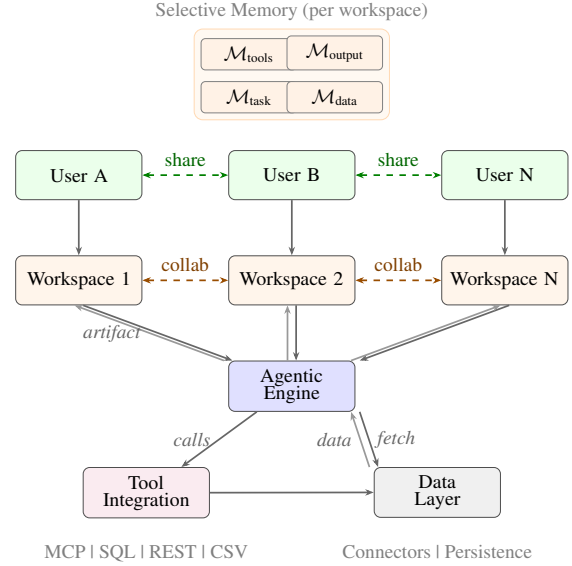
\begin{figure}[t]
\centering
\resizebox{0.95\columnwidth}{!}{%
\begin{tikzpicture}[
    node distance=0.9cm and 0.5cm,
    box/.style={
        rectangle, rounded corners=3pt, draw=black!70,
        fill=#1, minimum width=2.0cm, minimum height=0.8cm,
        text width=1.8cm, align=center, font=\small
    },
    box/.default={blue!8},
    membox/.style={
        rectangle, rounded corners=2pt, draw=black!50,
        fill=orange!10, minimum width=1.4cm, minimum height=0.5cm,
        text width=1.3cm, align=center, font=\footnotesize
    },
    arr/.style={-{Stealth[length=4pt]}, thick, black!60},
    lbl/.style={font=\footnotesize\itshape, text=black!60},
]

\node[box=green!8] (u1) {User A};
\node[box=green!8, right=1.4cm of u1] (u2) {User B};
\node[box=green!8, right=1.4cm of u2] (u3) {User N};

\draw[{Stealth[length=4pt]}-{Stealth[length=4pt]}, thick, green!50!black, dashed]
    (u1) -- node[above, font=\footnotesize, text=green!40!black] {share} (u2);
\draw[{Stealth[length=4pt]}-{Stealth[length=4pt]}, thick, green!50!black, dashed]
    (u2) -- node[above, font=\footnotesize, text=green!40!black] {share} (u3);

\node[box=orange!8, below=of u1] (w1) {Workspace 1};
\node[box=orange!8, below=of u2] (w2) {Workspace 2};
\node[box=orange!8, below=of u3] (w3) {Workspace N};

\draw[arr] (u1) -- (w1);
\draw[arr] (u2) -- (w2);
\draw[arr] (u3) -- (w3);

\draw[{Stealth[length=4pt]}-{Stealth[length=4pt]}, thick, orange!60!black, dashed]
    (w1) -- node[above, font=\footnotesize, text=orange!50!black] {collab} (w2);
\draw[{Stealth[length=4pt]}-{Stealth[length=4pt]}, thick, orange!60!black, dashed]
    (w2) -- node[above, font=\footnotesize, text=orange!50!black] {collab} (w3);

\node[membox, above=0.6cm of u2, xshift=-0.7cm] (m1) {$\mathcal{M}_\text{task}$};
\node[membox, above=0.6cm of u2, xshift=0.7cm]  (m2) {$\mathcal{M}_\text{data}$};
\node[membox, above=1.3cm of u2, xshift=-0.7cm] (m3) {$\mathcal{M}_\text{tools}$};
\node[membox, above=1.3cm of u2, xshift=0.7cm]  (m4) {$\mathcal{M}_\text{output}$};

\begin{scope}[on background layer]
    \node[fit=(m1)(m2)(m3)(m4), fill=orange!5, draw=orange!30,
          rounded corners=4pt, inner sep=3pt,
          label={[font=\footnotesize, text=black!50]above:Selective Memory (per workspace)}] {};
\end{scope}

\node[box=blue!12, below=of w2] (engine) {Agentic\\[-2pt]Engine};

\draw[arr] ([xshift=2pt]w1.south) -- ([xshift=2pt]engine.north west);
\draw[arr] ([xshift=2pt]w2.south) -- ([xshift=2pt]engine.north);
\draw[arr] ([xshift=2pt]w3.south) -- ([xshift=2pt]engine.north east);

\draw[arr, black!40] ([xshift=-2pt]engine.north west) -- node[left, lbl] {artifact} ([xshift=-2pt]w1.south);
\draw[arr, black!40] ([xshift=-2pt]engine.north) -- ([xshift=-2pt]w2.south);
\draw[arr, black!40] ([xshift=-2pt]engine.north east) -- ([xshift=-2pt]w3.south);

\node[box=purple!8, below left=0.9cm and 0.3cm of engine] (tools) {Tool\\[-2pt]Integration};
\node[box=gray!12, below right=0.9cm and 0.3cm of engine] (data) {Data\\[-2pt]Layer};

\draw[arr] (engine) -- node[left, lbl] {calls} (tools);
\draw[arr] ([xshift=2pt]engine.south east) -- node[right, lbl] {fetch} ([xshift=2pt]data.north west);
\draw[arr] (tools.east) -- (data.west);
\draw[arr, black!40] ([xshift=-2pt]data.north west) -- node[left, lbl] {data} ([xshift=-2pt]engine.south east);

\node[font=\footnotesize, text=black!50, below=0.3cm of tools] {MCP \textbar\ SQL \textbar\ REST \textbar\ CSV};
\node[font=\footnotesize, text=black!50, below=0.3cm of data] {Connectors \textbar\ Persistence};

\end{tikzpicture}%
}%
\caption{System architecture. Multiple users each own independent workspaces that can be shared for collaboration with role-based access control. Each workspace persists four categories of selective memory (top). The agentic engine composes workspace memory into session prompts; the zero-token data architecture decouples generated artifacts from runtime data. Data is accessed via heterogeneous connectors (CSV, SQL, REST, MCP).}
\label{fig:architecture}
\end{figure}

Figure~\ref{fig:architecture} presents the end-to-end architecture. Users interact with a conversational frontend; each session is contextualized by selective persistent memory drawn from the workspace store. The agentic engine generates and edits versioned artifacts through autonomous tool use, with external data accessed via a multi-connector integration layer (CSV, SQL, REST, MCP). Generated artifacts consume data exclusively through a runtime injection contract (Section~\ref{sec:zero-token}), and the workspace captures only the four structured memory categories---never the session trace.

\subsection{Problem Formulation}

Consider an agentic LLM system that, given a user query $q$ and system prompt $s$, produces an artifact $a$ through a sequence of tool-use steps $T = (t_1, t_2, \ldots, t_n)$. In current systems, the complete session state $\mathcal{S} = (s, q, T, a)$ is discarded after the session ends.

When the user returns for a related task $q'$, they must re-specify all implicit knowledge: the domain constraints encoded in $s$, the data schema assumptions in $q$, the tool configurations invoked during $T$, and the output format contracts in $a$. We observe that this re-specification is both the primary source of user friction and the primary source of wasted tokens.

\subsection{Memory Decomposition}

We decompose the reusable knowledge from a session into four categories:

\paragraph{Task Specifications ($\mathcal{M}_\text{task}$).} Custom instructions that extend the base system prompt. These encode domain rules (``use gate-based color coding: Supply blue, Orders orange, Process purple''), output preferences (``always include an executive summary card''), and quality constraints (``guard against NaN in all numeric operations''). Task specifications are authored by users and refined over multiple sessions.

\paragraph{Data Schemas ($\mathcal{M}_\text{data}$).} Precomputed summaries of associated data sources: column names and types, statistical distributions for numeric columns, unique value catalogs for categorical columns, row counts, and sample rows. These are generated automatically from uploaded data via statistical profiling (we use \texttt{pandas.describe()} with extensions for categorical analysis). Profiling reads the full table, so its cost scales with row count. What is bounded is the \emph{output}, and only along one axis: the summary is near-constant in the number of rows but grows linearly in the number of columns, since the profiler emits a per-column entry (Table~\ref{tab:tokens}). Data schemas provide the LLM with sufficient information to generate correct data-processing code without reading raw data---reducing prompt tokens while improving generation accuracy through structured metadata.

\paragraph{Tool Configurations ($\mathcal{M}_\text{tools}$).} The set of available external tools and data connectors with their parameter schemas, invocation patterns (REST endpoint, MCP server, SQL connection, sub-agent delegation), and authentication requirements. In enterprise settings, tool availability, connector configurations, and auth context are stable across sessions but expensive to discover and configure from scratch.

\paragraph{Output Constraints ($\mathcal{M}_\text{output}$).} Structural contracts between the generated artifact and its runtime environment. The critical constraint is the data-injection contract: generated scripts must consume data exclusively from a well-defined runtime injection point, never from hardcoded values. This constraint enables zero-token data refresh (Section~\ref{sec:zero-token}).

\subsection{Selective Forgetting}

Equally important is what we \emph{do not} persist. Given a session trace $T = (t_1, t_2, \ldots, t_n)$, we discard:

\begin{itemize}
    \item \textbf{Intermediate temporary files:} Working files created during generation (partial drafts, intermediate versions, scratch computations) that are artifacts of one execution path.
    \item \textbf{Unapproved changes:} Modifications the user explored but did not explicitly save. The undo stack (Section~\ref{sec:implementation}) allows users to try refinements risk-free; only changes the user accepts are persisted.
    \item \textbf{Reasoning traces:} The LLM's intermediate chain-of-thought and planning steps, which reflect session-specific problem decomposition.
    \item \textbf{Tool invocation logs:} The specific file reads, writes, and shell commands executed during generation, which reflect one solution path among many.
    \item \textbf{Error recovery paths:} Failed tool calls, retries, and workarounds, which are artifacts of a specific session's execution.
    \item \textbf{Raw data:} The complete source data, which is replaced by compact schema summaries.
\end{itemize}

This selective forgetting is loosely motivated by the ``lost in the middle'' finding \citep{liu2024lost}, but our claim is about \emph{content} rather than position: a prior session's tool trace is not merely inert but potentially misleading, biasing the agent toward repeating a previous solution path.

\subsection{Memory Composition at Session Start}

When a new session begins, the system prompt is composed from persistent memory:
\begin{equation}
    s' = s_\text{base} \oplus \mathcal{M}_\text{task} \oplus \mathcal{M}_\text{tools} \oplus \mathcal{M}_\text{output},
\end{equation}
where $s_\text{base}$ is the default system prompt and $\oplus$ denotes structured concatenation with section headers. The data schema $\mathcal{M}_\text{data}$ is injected into the user message alongside the query:
\begin{equation}
    q' = q \oplus \mathcal{M}_\text{data}.
\end{equation}

This composition is performed at the application layer before the LLM is invoked. The LLM receives a fully contextualized prompt without any awareness of the persistence mechanism.

\subsection{Zero-Token Data Refresh}
\label{sec:zero-token}

A complementary mechanism to selective memory is the \textbf{data-injection contract}: generated artifacts must consume data exclusively from a runtime injection point, never from hardcoded values. The LLM generates programs that parse, aggregate, and render dynamically---including insight text (summaries, outliers, rankings)---so that when the underlying data changes, the artifact re-renders correctly without any LLM re-invocation.

When new data arrives (from any connector type), the system checks schema compatibility: the original column set must be a subset of the new source's columns. If compatible, data refreshes with zero tokens. If the schema has diverged, the user is prompted to regenerate. This enables additive schema evolution (new columns are permitted; missing columns trigger regeneration).

\section{Collaborative Memory Sharing}
\label{sec:sharing}

A workspace $\mathcal{W}$ encapsulates the full selective memory state:
\begin{equation}
    \mathcal{W} = (\mathcal{M}_\text{task}, \mathcal{M}_\text{data}, \mathcal{M}_\text{tools}, \mathcal{M}_\text{output}, a, V),
\end{equation}
where $a$ is the most recent generated artifact and $V$ is its version history. The workspace stores structured configuration, generated code, and version metadata, never conversational transcripts, reasoning traces, or tool invocation logs. A saved workspace is a curated artifact store, not a chat history.

The sharing workflow exploits the zero-token data architecture:
\begin{enumerate}
    \item \textbf{Load:} A colleague loads the shared workspace, restoring all selective memory into their session. Draft isolation ensures their edits do not affect the published artifact.
    \item \textbf{Swap:} They connect their own data source (e.g., their region's SQL database, a different REST endpoint, or an uploaded CSV). If schema-compatible, the artifact renders with new data---zero tokens.
    \item \textbf{Refine:} They ask refinement questions within the established context (e.g., ``add a trend chart''). The LLM benefits from the persisted task specifications and tool configurations without re-specification.
    \item \textbf{Query:} Viewers without edit access can use the AI query mode to ask questions about the artifact and its underlying data, receiving answers contextualized by the workspace's selective memory.
\end{enumerate}

Access is managed through OIDC-based identity with three roles: \emph{owners} have full control over the workspace and its memory; \emph{stewards} can edit artifacts and modify task specifications; \emph{viewers} can load, query, and extend with their own data but cannot modify the base workspace.

\section{Implementation}
\label{sec:implementation}

We implement shared selective persistent memory in a deployed collaborative workspace platform consisting of a FastAPI backend, an agentic LLM engine, a multi-connector data layer, and a conversational frontend.

\paragraph{Agentic Loop.} Each user turn initializes a fresh session with the composed system prompt (Equation 1) and data summary (Equation 2). The agent uses Claude Opus 4 \citep{anthropic-claude-2025} via the Claude Agent SDK \citep{anthropic-agent-sdk-2025} with autonomous tool use (file I/O, code search, shell execution), scoped to the user's session workspace. Two interaction modes are supported: \emph{edit mode} for artifact generation and modification, and \emph{query mode} for read-only questions over existing artifacts.

\paragraph{Data Integration.} A unified connector interface abstracts four source types---CSV upload, SQL databases, REST APIs, and MCP servers (JSON-RPC with mTLS)---into a normalized tabular representation. Rather than injecting raw data into the LLM prompt, the system precomputes a compact statistical profile (column types, distributions, categorical catalogs, sample rows) that captures the information needed for correct code generation. On a typical enterprise table this costs ${\sim}$500 tokens, a 100$\times$ reduction over raw injection; on wide tables it is larger, since profile size grows with column count (Section~\ref{sec:evaluation}).

\paragraph{Persistence and Versioning.} Workspace metadata---the four memory categories, access control lists, and data source configurations---is persisted to MongoDB. Generated artifacts and their version history are managed through git, providing full version control with diff, branch, and rollback capabilities. No conversation history or reasoning traces are stored in either layer. Users edit within isolated draft sessions; in-progress modifications do not affect the published artifact until explicitly committed via a publish operation that creates a new git-versioned snapshot.

\paragraph{Save and Revert.} An in-session undo stack (up to 10 snapshots) enables risk-free exploration---users can revert to any prior version without re-invoking the model. Only explicitly saved changes persist, reinforcing selective forgetting.

\section{Evaluation}
\label{sec:evaluation}

We evaluate shared selective persistent memory through four studies: a controlled ablation comparing memory conditions, a public dataset replication for reproducibility, a token efficiency analysis, and a user study.

\subsection{Experimental Setup}

All experiments use the deployed system described in Section~\ref{sec:implementation}. The agentic engine uses Claude Opus 4 (\texttt{claude-opus-4-6}) \citep{anthropic-claude-2025} for primary artifact generation and Claude Sonnet 4 for lightweight subtasks (schema validation, tool dispatch), accessed through the Claude Agent SDK at default sampling settings. Every cell is a single run, so we report no dispersion. We evaluate on a corpus of 24 real enterprise data files spanning supply chain operations, sales reporting, and process metrics, with sizes ranging from 200 rows / 8 columns to 45K rows / 42 columns. Data is ingested via CSV upload and SQL connectors.

\paragraph{Task definition.} Each experimental task requires generating an interactive artifact from a structured data source with specific formatting requirements (color thresholds, layout preferences, summary cards). There is one task per data file, so the 24 files and the 24 tasks are in one-to-one correspondence. Tasks and their formatting requirements were authored by the authors, drawn from artifacts previously requested by internal users of the deployed system; they were not written by independent annotators, and we note this as a degree of freedom in Limitations. Tasks are drawn from the three use case families described in Section~\ref{sec:usecases}: 10 tasks model recurring artifact refresh (Use Case~1), 8 tasks model cross-team workspace adaptation (Use Case~2), and 6 tasks model iterative artifact construction (Use Case~3). We define two task types: \emph{initial generation} (first artifact from a new data source) and \emph{recurring generation} (same task structure, new data).

\subsection{Experiment 1: Memory Condition Ablation}

We compare three conditions across 24 recurring generation tasks: \emph{no memory} (default prompt, full re-specification), \emph{full history} (complete prior session transcript injected), and \emph{shared selective memory} (four structured categories plus artifact; traces discarded).

\paragraph{Protocol.} Each artifact was evaluated by two raters on four pass/fail criteria: render correctness, data fidelity (spot-checked against source), format compliance, and completeness. A task passed only if both raters scored pass on all four ($\kappa = 0.91$). Both raters were colleagues from a business team who took no part in the system's design, implementation, or evaluation, and neither is an author. They saw the rendered artifact and task description in randomized order; condition labels, transcripts, and token counts were withheld. One limit: format compliance is one of the four criteria and only some conditions carry a format specification, so a rater seeing an artifact that ignores it can plausibly infer the condition.

\begin{table}[t]
\centering
\small
\begin{tabular}{@{}lccc@{}}
\toprule
\textbf{Metric} & \textbf{No Mem.} & \textbf{Full Hist.} & \textbf{Selective} \\
\midrule
Input tokens (K) & 2.1 & 18.7 & 3.4 \\
Output tokens (K) & 8.2 & 9.6 & 4.1 \\
User turns & 4.3 & 3.1 & 1.4 \\
Completion (\%) & 79 & 71 & \textbf{96} \\
Time (sec) & 285 & 310 & 68 \\
\bottomrule
\end{tabular}
\caption{Ablation results across 24 recurring artifact generation tasks (means). Shared selective memory achieves the highest completion rate with the fewest tokens and turns. Full history increases input tokens 9$\times$ over no-memory but \emph{decreases} completion rate.}
\label{tab:ablation}
\end{table}

\paragraph{Results (Table~\ref{tab:ablation}).} Selective memory completes 23/24 tasks (96\%) in 1.4 turns, vs.\ 19/24 (79\%) in 4.3 turns for no-memory and 17/24 (71\%) in 3.1 turns for full history. Full history completes fewer tasks than no-memory despite 9$\times$ more input tokens. By use case, as counts: recurring refresh (UC1, $n=10$) 10/10 vs.\ 7/10 vs.\ 6/10; cross-team adaptation (UC2, $n=8$) 8/8 vs.\ 7/8 vs.\ 6/8; iterative construction (UC3, $n=6$) 5/6 under all three conditions, though selective memory halves the turns (2.8 vs.\ 6.2). Output tokens drop 50\% (4.1K vs.\ 8.2K) and mean wall-clock time drops from 285s to 68s.

\paragraph{Two mechanisms are aggregated in the selective column.} For 18 of the 24 tasks zero-token refresh fired and the model was never invoked; the remaining 6 invoked it with composed memory. The selective column of Table~\ref{tab:ablation} therefore averages artifact reuse with memory-conditioned generation, so its token and latency figures are not per-invocation costs. Completion is unaffected---both paths use the same rubric---and per-task records were not retained, so we report the confound rather than the split.

\paragraph{Significance.} The design is paired---the same 24 tasks are run under all three conditions---so the appropriate test is McNemar's rather than Fisher's exact test. Task-level pairings were not retained during the original run, so we report bounds over all pairings consistent with the observed marginals (23/24 selective, 19/24 no-memory, 17/24 full history). Under the most favorable admissible pairing, selective vs.\ no-memory gives exact McNemar $p = 0.125$ and selective vs.\ full history $p = 0.031$; under the least favorable, $p = 0.219$ and $p = 0.070$ respectively. Applying a Bonferroni correction across the three pairwise comparisons ($\alpha = 0.0167$), neither comparison reaches significance at $n = 24$ under any pairing.

We therefore report the completion-rate differences as descriptive rather than inferential: the direction is consistent across all three use-case families, but 24 paired tasks cannot establish it.

\paragraph{Failures.} The single selective-memory failure involved a task requiring a join across two uploaded files. $\mathcal{M}_\text{data}$ profiles each source independently and records no cross-source key relationships, so the composed prompt described both schemas without expressing how they related. This is a structural limit of schema-summary-as-memory: the category carrying data structure has no representation for relationships \emph{between} sources.
No-memory failures (5/24) were all format non-compliance. Full-history failures (7/24) were dominated by the agent reproducing a solution path from the injected transcript rather than responding to the current request. This was not annotated against a pre-registered protocol, so we report it as an observation rather than a measured cause (Limitations).

\subsection{Experiment 2: Public Dataset Replication}
\label{sec:public}

To enable reproduction outside the proprietary enterprise corpus, we replicate the ablation on four datasets retrieved from their published sources, spanning a wide range of table widths: World Bank GDP (13{,}979 rows, 4 cols), UCI Adult Income (32{,}561 rows, 15 cols), Global Superstore orders (1{,}000-row published sample, 24 cols), and NYC 311 Service Requests (5{,}000 rows, 44 cols).\footnote{Sources: Kaggle/plotly (Global Superstore), UCI ML Repository (Adult), World Bank via Frictionless Data (GDP), NYC Open Data (311). Trial-level outcomes are released with the artifact.}

Note that no-memory completion is not comparable across Tables~\ref{tab:ablation} and~\ref{tab:public-ablation}. In the enterprise ablation the user re-specifies the constraints, taking 4.3 turns, so that condition reaches 79\% without needing memory at all. Here the specification is withheld at Phase~2, so a condition either carries it or fails. The two experiments measure different things: one the turns memory saves, the other whether the task is achievable without it.

\paragraph{Protocol.} Every condition runs an identical Phase~1: generate an artifact from V1 data given the full task prompt \emph{and} the formatting specification, establishing the constraint. Every condition then receives an identical Phase~2 instruction---``Regenerate the dashboard with this updated data.''---over schema-compatible V2 data (same columns, rows shuffled, numerics perturbed $\pm$15\%). The specification is \emph{not} restated. Conditions differ only in what they carry into Phase~2: nothing (no memory), the Phase~1 transcript including its artifact (full history), or $\mathcal{M}_\text{task}$ (selective). Three runs per dataset per condition, 36 trials, 72 model calls.

Completion requires all validation criteria to pass, matching the all-criteria rule used by our human raters. Alongside the structural checks (valid render, data-injection contract, no hardcoded data, parser present, correct column references) we add \emph{spec compliance}: the specification names a background colour that appears in no task prompt, so an artifact exhibits it only if the specification reached the session. This is the criterion that discriminates; the structural checks are satisfied by any working pipeline.

\begin{table}[t]
\centering
\small
\begin{tabular}{@{}lccc@{}}
\toprule
\textbf{Metric} & \textbf{No Mem.} & \textbf{Full Hist.} & \textbf{Selective} \\
\midrule
Completion & 0/12 & 8/12 & \textbf{12/12} \\
Spec compliance & 0/12 & 8/12 & \textbf{12/12} \\
Input tokens (K) & 3.8 & 7.7 & \textbf{3.9} \\
Output tokens (K) & 5.5 & 3.5 & 4.3 \\
Time, median (s) & 130 & \textbf{99} & 140 \\
Tool calls & 3.7 & 3.6 & 10.1 \\
\bottomrule
\end{tabular}
\caption{Public dataset ablation: 4 datasets $\times$ 3 conditions $\times$ 3 runs. Selective memory completes every trial at essentially the same input cost as no memory (3.9K vs.\ 3.8K), which completes none; full history reaches 8/12 at twice the input tokens. Times are medians; two trials exceeded 600s and distort the means.}
\label{tab:public-ablation}
\end{table}

\paragraph{Results (Table~\ref{tab:public-ablation}).} The separation is consistent across all four datasets: no memory completes 0/3 on every one, selective memory 3/3 on every one, and full history varies between 1/3 and 3/3. Completion tracks spec compliance exactly---every no-memory failure is a spec failure and nothing else. Those artifacts are structurally sound, scoring 7 of 8 criteria: valid, parseable, contract-following. They simply do not reproduce a specification they were never given.

What this isolates is the workspace, not one component of it. A workspace is $(\mathcal{M}_\text{task}, \mathcal{M}_\text{data}, \mathcal{M}_\text{tools}, \mathcal{M}_\text{output}, a, V)$ (Section~\ref{sec:sharing}), so the selective condition resumes Phase~1 with both the specification and the prior artifact $a$, while the others begin empty. Either could carry the constraint; this experiment does not separate them. The same asymmetry explains its higher latency and tool-call count: it revises an existing artifact where the others write a new one.

The cost comparison is the stronger result. Counting input and output together, selective memory is the \emph{cheapest} of the three at 8.2K tokens per task, against 9.3K for no memory and 11.2K for full history: the condition that completes every trial costs less than the one that completes none, which spends 5.5K output tokens producing artifacts that fail. Persisting the specification is not merely close to free, it is cheaper than omitting it. Full history reaches 8/12 for 7.7K---twice the input budget---so a raw transcript does carry the specification, just at double the price and less reliably. All four of its failures produced no artifact at all rather than a non-compliant one, and they span three of the four datasets. Transcript length did not predict them: the shortest transcript failed while longer ones succeeded. We therefore record this as instability under transcript injection rather than as evidence of trace anchoring, which would predict a non-compliant artifact rather than none.

\paragraph{Significance.} The design is paired and per-trial outcomes were retained, so exact McNemar applies directly. Selective vs.\ no memory gives $p = 0.0005$ and full history vs.\ no memory $p = 0.008$; both clear a Bonferroni-corrected $\alpha = 0.0167$ across three comparisons. Selective vs.\ full history does not ($p = 0.125$). We are careful about how to read that. The observed gap is large---12/12 against 8/12, a 33-point difference in completion---and it runs the same way on three of the four datasets, with a single consistent failure mode. What the test says is not that the conditions perform alike, but that four discordant pairs cannot establish a difference of that size: at full history's observed 33\% failure rate, roughly 21 trials per condition would be needed. We therefore report the gap as observed but unestablished, and rest the claim that structured memory is preferable on the cost comparison, which is unambiguous at this sample.

\paragraph{Zero-token refresh.} Evaluated separately from memory composition, since conflating them is what made the original table uninterpretable. Across the same 36 trials the refresh path fired on 12/12 selective trials with no model call, at a median 0.08s. Because V2 preserves V1's columns by construction, the compatibility check is guaranteed to pass; this measures that the mechanism executes and that artifacts written under the data-injection contract survive a data swap, not how often real updates are schema-compatible.

This qualifies the latency column of Table~\ref{tab:public-ablation}: in the ablation the selective condition regenerates after the refresh so all three perform the same generation, whereas in deployment it stops at the refresh---0.08s with no model call, against the 140s we report. The 140s is the price of the controlled comparison, not what a user pays.
\subsection{Experiment 3: Token Efficiency}
\label{sec:tokens}

We measure the token cost of data representation comparing three strategies: (1)~raw data injected verbatim, (2)~truncated data (first 50 rows), and (3)~summary-driven (statistical profile from \texttt{summarize\_data()}).

\begin{table}[t]
\centering
\small
\begin{tabular}{@{}lrrr@{}}
\toprule
\textbf{Data size} & \textbf{Raw} & \textbf{Trunc.} & \textbf{Summary} \\
\midrule
Small ($<$1K rows) & 3.2K & 1.8K & 0.4K \\
Medium (1--10K) & 28.5K & 1.9K & 0.5K \\
Large ($>$10K) & 142.3K & 2.0K & 0.6K \\
\midrule
\textbf{Mean (enterprise)} & \textbf{48.7K} & \textbf{1.9K} & \textbf{0.5K} \\
\textbf{Mean (public)} & \textbf{789.1K} & \textbf{4.4K} & \textbf{1.8K} \\
\bottomrule
\end{tabular}
\caption{Data representation tokens by strategy. Summary size is bounded in \emph{rows} but scales with \emph{columns}: the enterprise buckets hold near 0.5K as row counts grow three orders of magnitude, while the 44-column NYC 311 table costs 4.5K. Reduction versus raw injection is 97$\times$ on enterprise data and 431$\times$ on the public set; versus truncation---the strategy a practitioner would actually reach for---it is 2--3$\times$.}
\label{tab:tokens}
\end{table}

\paragraph{Results (Table~\ref{tab:tokens}).} Summary-driven generation consumes a mean of 0.5K tokens against 48.7K for raw injection on enterprise data (97$\times$) and 1.8K against 789.1K on the public datasets (431$\times$).

\paragraph{The bound is on rows, not columns.} Our submitted version claimed sub-1K summaries regardless of dataset size; re-running on the public data shows that holds only for row count. Summary size is essentially flat in rows---the enterprise buckets move from 0.4K to 0.6K as rows grow three orders of magnitude, and UCI Adult's 32{,}561 rows cost 922 tokens---but scales monotonically with columns, since the profiler emits a per-column entry and enumerates categorical values. Across the four public datasets, ordered by width, summaries run 339, 922, 1{,}584 and 4{,}486 tokens for 4, 15, 24 and 44 columns. Row count does not track this at all: Superstore has 1{,}000 rows and a larger summary than UCI Adult's 32{,}561. No dataset in our original set exceeded 15 columns, so this axis went untested. Summarization decouples prompt cost from \emph{table length}, the dimension that varies across refreshes of one source; cost in \emph{table width} remains linear and binds on wide tables.

\paragraph{Against truncation.} The informative baseline is not raw injection but truncation, and there the reduction is a modest 2--3$\times$ (1.9K$\rightarrow$0.5K enterprise, 4.4K$\rightarrow$1.8K public), narrowing to 2.05$\times$ on the widest table. Summaries matter less for the token saving than for fidelity: truncation to the first 50 rows discards tail distributions and rare categories, causing errors in 5 of 24 enterprise tasks, while the statistical profile preserves them.

\subsection{Experiment 4: User Study}
\label{sec:user-study}

A user study ($N=12$; 6 engineers, 6 analysts) across four counterbalanced tasks showed recurring generation was 14$\times$ faster with shared selective memory (12s vs.\ 165s, on the zero-token refresh path; the 4.2$\times$ in Table~\ref{tab:ablation} is the ablation mean over both paths), refinement 2.5$\times$ faster, and constrained generation 3$\times$ faster. All Likert dimensions favored selective memory, with the largest gap on ``would use again'' (6.5 vs.\ 4.2 / 7). Participants used revert 1.8 times per session on average.

\subsection{Summary of Findings}

Table~\ref{tab:summary} in Appendix~\ref{sec:appendix-extra} collects the measured effects, restricted to quantities an experiment in this paper estimates.

\section{Discussion and Conclusion}
\label{sec:conclusion}

Our two ablations appear to disagree about full history, and the disagreement is informative. On enterprise tasks it completes no more than no memory (17/24 vs.\ 19/24) at 9$\times$ the input; on public data it completes 8/12 against 0/12, significantly. One design difference explains both. The enterprise no-memory condition \emph{re-specifies}, taking 4.3 turns against 1.4, and a condition given full re-specification does not need memory to succeed---so memory buys turns, not completions. The public design withholds the specification, where a condition either carries it or fails. Persisted context does not improve what an agent can do when the user will say everything again; it removes the need to.

Full history supplies that context too, at twice the token cost and with four unexplained failures in twelve. The decomposition is not specific to any artifact type; it applies to dashboards, reports, documents, and any workspace where an agent produces artifacts from structured data.

Quantitatively, selective memory completes 23/24 enterprise tasks against 19/24 with no memory and 17/24 with full history, though at $n = 24$ no pairwise difference survives correction for multiple comparisons (Section~\ref{sec:evaluation}); that comparison motivates a powered replication rather than settling anything. The public replication is the stronger evidence: 12/12 against 0/12 and 8/12, with both differences from no memory surviving Bonferroni correction, and every no-memory failure attributable to the specification criterion alone. The two efficiency results are architectural rather than statistical: zero-token refresh removes the model call entirely for schema-compatible updates, firing on 12/12 public trials at a median 0.08s and measured in the user study as a 14$\times$ task-time reduction on that path, and summary-driven generation costs 97--431$\times$ fewer tokens than raw injection when the model is invoked. Both reduce how often and how expensively the LLM is called rather than improving model reasoning. Git-backed versioning and revert give users agency over the memory lifecycle, enabling risk-free exploration and restoration to any prior state without re-invoking the model.

Our results suggest that context management, not model capability, is the primary lever for improving agentic LLM efficiency in enterprise settings.

\section*{Limitations}

The current memory decomposition is manually designed; while it generalizes across the artifact types evaluated (dashboards, reports, documents), automatically identifying which context elements are reusable across sessions and artifact types remains an open problem. The zero-token data architecture is limited to structured tabular data with stable schemas; extending to streaming data, unstructured documents, or real-time API responses would require richer schema compatibility checking. While the multi-connector layer supports CSV, SQL, REST, and MCP sources, each connector type introduces its own failure modes (connection timeouts, authentication expiry, schema drift) that are handled independently rather than through a unified retry strategy. Our four-criterion rubric captures functional correctness and format compliance but does not assess subjective qualities such as visual aesthetics or information hierarchy. Our user study ($N=12$) is sufficient for identifying trends but underpowered for statistical significance testing on individual Likert items; a larger study is warranted. Collaboration is workspace-level; finer-grained sharing of individual memory components (e.g., sharing tool configurations without task specifications) is not yet supported.

\paragraph{Evidence.} Several limits attach to the evidence rather than the architecture. Every cell is a single run with no repeated seeds, and on the enterprise tasks no completion difference survives correction for multiple comparisons. Three baselines are absent: a summarized-history condition, a hand-written instructions file (the comparison a practitioner would make), and a leave-one-component-out ablation. Full history is also confounded, differing from selective memory both in what information is present and in how well it is written. All experiments use a single model family, with no head-to-head comparison against MemGPT, A-MEM, Mem0, or RAG. The public sample is underpowered for the comparison we most want: 12/12 against 8/12 needs roughly 21 trials per condition, not 12. Its schema-compatibility rate is also 100\% by construction rather than by measurement. Finally, tasks were authored by the authors and scored against a rubric the authors defined; the raters were independent, but a replication with third-party task authors would be stronger evidence than anything reported here.

\paragraph{Future Work.}
The most valuable next step is the evaluation this paper does not contain: a powered, paired replication with repeated runs per cell, a summarized-history condition, a hand-written instructions-file baseline, and a leave-one-component-out ablation of the four categories. Beyond that, three directions follow from the architecture. \emph{Learned memory selection} would replace our fixed decomposition with a classifier over session traces. \emph{Memory decay} would age out specifications not validated against recent sessions, which participants asked for in the user study. \emph{Agent-level memory} would retain effective tool sequences and error-recovery heuristics rather than workspace configuration alone; it differs from workspace memory in kind, being learned and private to the agent rather than user-authored and shared, so it would need generalization boundaries, staleness detection, and a conflict-resolution rule against explicit specifications---a pattern that works in one data domain need not transfer to another. Extending the connector layer to streaming and event-driven refresh would broaden applicability beyond batch workflows.

\section*{Ethical Considerations}

The system does not store or process personally identifiable information (PII). All persistent memory operates at the workspace level---task specifications, data schemas, tool configurations, and output constraints---none of which contain user-level personal data. The data sources used in our evaluation consist entirely of aggregated operational metrics. Personalization is scoped to workspace configuration, not to individual user profiles or behavioral tracking. The multi-connector architecture introduces additional security considerations: SQL connectors require credential management, REST connectors handle API tokens, and MCP connectors use mTLS authentication---all scoped to per-workspace configuration with no cross-workspace credential leakage. The system generates executable code from natural language instructions; file system access is scoped to per-user session directories and generated artifacts are served in sandboxed browser contexts. A generated artifact receives data only through the runtime injection point and holds no connector handles or credentials, so it cannot open a connection of its own: connector calls are resolved in the backend against the requesting session's workspace configuration and rejected if the connector is not attached to that workspace.

\section*{Acknowledgements}

We used Claude (Anthropic) as an AI writing assistant during manuscript preparation, including drafting, editing, and LaTeX formatting. All intellectual contributions---system design, architecture decisions, implementation, experimental design, and analysis---are the authors' own. The authors reviewed and take full responsibility for all content.


\bibliography{custom}

\appendix

\section{Memory Composition Example}
\label{sec:appendix-memory}

The following illustrates how a composed system prompt is constructed from selective memory at session start. The base system prompt ($s_\text{base}$) contains domain-agnostic generation instructions---data architecture rules, output size constraints, and code patterns. Task specifications ($\mathcal{M}_\text{task}$) are appended as user-authored ``Additional Instructions.'' Tool configurations ($\mathcal{M}_\text{tools}$) are appended with per-tool invocation templates. The data schema ($\mathcal{M}_\text{data}$) is injected into the user message alongside the query.

\begin{lstlisting}[basicstyle=\scriptsize\ttfamily]
# Composed system prompt (s')

## Base system prompt (~500 tokens)  # s_base
You are an expert at creating interactive
dashboards and data-driven artifacts from
structured data sources.
CRITICAL: Data Architecture - Zero-Token Refresh
- Data is provided via a runtime injection
  point -- parse dynamically, never hardcode
- ALL aggregation, filtering, totals computed
  from parsed data at runtime
- Insight text must be generated dynamically
  (e.g., top performers, outliers)
- Guard against NaN/undefined in all numeric
  operations: always coerce with fallback to 0
- Data may come from CSV, SQL, REST, or MCP
  connectors -- treat all sources uniformly

## Additional Instructions             # M_task
Always use gate-based color coding:
  Supply=#3B82F6, Orders=#F97316,
  Process=#8B5CF6, Exec=#10B981
Include executive summary cards with KPIs.
Use dark theme (#1a1a2e background).
Format percentages to 1 decimal place.
Show attainment >= 95% in green, < 80% red.

## Available External Tools             # M_tools
You can call external tools via the Bash tool.

### File a Bug
Description: File a bug report in the
  issue tracker
Parameters:
  - title: Bug title (required)
  - component: Component name (required)
  - description: Bug description (required)
To call:
  curl -s -X POST http://localhost:<port>/
    api/tools/execute
    -H "Content-Type: application/json"
    -d '{"tool":"file_bug",
         "params":{"title":"..."}}'

### Search Knowledge Base
Description: Search documentation
Parameters:
  - query: Search text (required)
  - limit: Max results (default 10)
To call:
  curl -s -X POST http://localhost:<port>/
    api/tools/execute
    -H "Content-Type: application/json"
    -d '{"tool":"search_kb",
         "params":{"query":"..."}}'

## Output Constraints                   # M_output
Generated scripts must consume data
exclusively from the runtime injection
point. Never embed data values in source.
Keep output under 50KB. Use JS functions
to render tables and charts -- never write
repetitive markup for each row.
\end{lstlisting}

\begin{lstlisting}[basicstyle=\scriptsize\ttfamily]
# User message (q')

[User query: "Generate a supply chain
  operations dashboard"]

## Data: weekly_ops.csv                 # M_data
(450 rows, 28 columns)
Columns: LOB, Gate, ExecToGo, Shipped,
  Target, Attainment, Region, Week, ...

Numeric columns summary (12 total):
         ExecToGo  Shipped  Target  Attainment
count      450.0   450.0    450.0     450.0
mean      2341.2  8923.4   9100.0      93.7
std        812.5  3201.1   2800.0      12.4
min        120.0   450.0    500.0      42.0
max       5600.0 18200.0  18000.0     112.0

Categorical columns (unique values):
- LOB: Product A, Product B, Product C, Services, Product D
- Gate: Supply, Orders, Process, Exec
- Region: Americas, EMEA, APAC, Japan

Sample rows (first 3):
LOB=Product A, Gate=Supply, ExecToGo=3200,
  Shipped=12400, Target=13000, Attain=95.4%
LOB=Product C, Gate=Orders, ExecToGo=1800,
  Shipped=6200, Target=7000, Attain=88.6%
...
\end{lstlisting}

\section{Supplementary Tables and Figures}
\label{sec:appendix-extra}

\begin{table}[t]
\centering
\small
\begin{tabular}{@{}lp{4.5cm}@{}}
\toprule
\textbf{Mechanism} & \textbf{Measured Effect} \\
\midrule
Specification memory & Public: 12/12 vs.\ 0/12 without memory ($p{=}0.0005$), at equal input cost. Enterprise: 3$\times$ fewer turns \\
Selective vs.\ full hist. & Half the input tokens (3.9K vs.\ 7.7K); 12/12 vs.\ 8/12 completion, observed but n.s.\ at $n{=}12$ \\
Zero-token refresh & 12/12 refresh and validate; 0 LLM tokens; 0.08s median latency \\
Summary-driven gen. & 97--431$\times$ vs.\ raw injection; 2--3$\times$ vs.\ truncation \\
\bottomrule
\end{tabular}
\caption{Measured effects, restricted to quantities an experiment in this paper actually estimates. Git-backed versioning with revert and cross-user workspace transfer are implemented and described in Sections~\ref{sec:sharing} and~\ref{sec:implementation}, but no study here evaluates them---no experiment varied the number of users sharing a workspace---so they are omitted rather than listed as findings. ``n.s.'' marks a difference the sample cannot resolve. Public figures are exact McNemar on paired per-trial outcomes; enterprise completion differences do not reach significance at $n{=}24$ and are omitted here.}
\label{tab:summary}
\end{table}

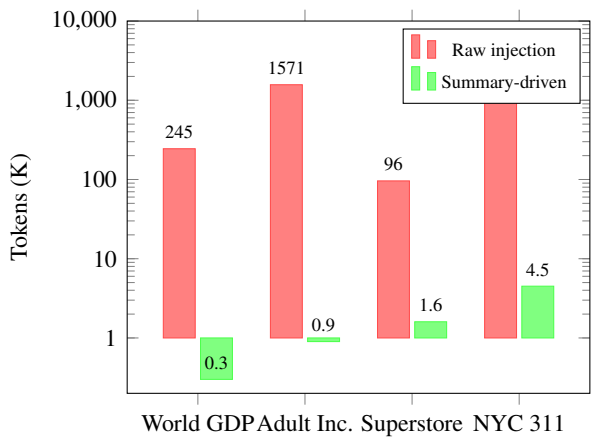
\begin{figure}[t]
\centering
\begin{tikzpicture}
\begin{axis}[
    ybar,
    width=\columnwidth,
    height=6.5cm,
    ymode=log,
    ylabel={Tokens (K)},
    ylabel style={font=\small},
    symbolic x coords={{World GDP}, {Adult Inc.}, {Superstore}, {NYC 311}},
    xtick=data,
    x tick label style={font=\small},
    y tick label style={font=\small},
    ytick={1, 10, 100, 1000, 10000},
    yticklabels={1, 10, 100, 1{,}000, 10{,}000},
    legend style={font=\scriptsize, at={(0.98,0.98)}, anchor=north east},
    bar width=12pt,
    ymin=0.2, ymax=10000,
    enlarge x limits=0.22,
    every node near coord/.append style={font=\scriptsize, anchor=south},
]
\addplot[fill=red!50, draw=red!70,
    point meta=explicit symbolic,
    nodes near coords
] coordinates {
    ({World GDP}, 244.5) [245]
    ({Adult Inc.}, 1570.8) [1571]
    ({Superstore}, 96.0) [96]
    ({NYC 311}, 1245.2) [1245]
};
\addplot[fill=green!50, draw=green!70,
    point meta=explicit symbolic,
    nodes near coords
] coordinates {
    ({World GDP}, 0.3) [0.3]
    ({Adult Inc.}, 0.9) [0.9]
    ({Superstore}, 1.6) [1.6]
    ({NYC 311}, 4.5) [4.5]
};
\legend{Raw injection, Summary-driven}
\end{axis}
\end{tikzpicture}
\caption{Token cost by strategy on the four public datasets (log scale), ordered by column count (4, 15, 24, 44). Raw injection tracks total table size; the summary tracks width alone, rising monotonically with columns (339, 922, 1{,}584, 4{,}486 tokens) while being indifferent to rows---Superstore has 1{,}000 rows and a larger summary than UCI Adult's 32{,}561. Summarization decouples prompt cost from table length, not from table width.}
\label{fig:token-reduction}
\end{figure}

\end{document}